\documentclass[11pt,a4paper]{article}
\usepackage[hyperref]{acl2020}
\usepackage{times}
\usepackage{latexsym}

\usepackage{microtype}

\usepackage{url}
\usepackage{graphicx}
\usepackage{booktabs}
\usepackage{amssymb,amsmath}
\usepackage{multirow}
\usepackage{subcaption}

\usepackage{comment}

\aclfinalcopy % Uncomment this line for the final submission
\def\writers#1#2{}

\title{GASP!\\ Generating Abstracts of Scientific Papers from Abstracts of Cited Papers}

\author{F.M.Zanzotto$^{(*)}$\\
  \\
  \And
  V.Bono$^{(\ddag)}$ \\
  $^{(*)}$
  University of Rome\\
  Tor Vergata \\
  Rome, Italy\\
  \And
  P.Vocca$^{(\star)}$ \\
  \\
  \And
  A.Santilli$^{(*)}$ \\
   $^{(\ddag)}$ University of Turin \\
  Turin, Italy
  \\
  \And
   D.Croce$^{(*)}$ \\
  \And
  G.Gambosi$^{(*)}$ \\
   $^{(\star)}$University of Tuscia \\
  Viterbo, Italy 
  \\
  \And
  R.Basili$^{(*)}$  \\
  \\}

\date{}

\begin{document}
\maketitle
\begin{abstract}
Creativity is one of the driving forces of human kind as it allows to break current understanding to envision new ideas, which may revolutionize entire fields of knowledge. Scientific research offers a challenging environment where to learn a model for the creative process. In fact, scientific research is a creative act in the formal settings of the scientific method and this creative act is described in articles.

In this paper, we dare to introduce the novel, scientifically and philosophically challenging task of Generating Abstracts of Scientific Papers from abstracts of cited papers (GASP) as a text-to-text task to investigate scientific creativity, To foster research in this novel, challenging task, we prepared a dataset by using services where that solve the problem of copyright and, hence, the dataset is public available with its standard split.  Finally, we experimented with two vanilla summarization systems to start the analysis of the complexity of the GASP task.

\end{abstract}

\section{Introduction}
%\writers{Fabio}{Viviana, Paola}

%VIVIANA TESTO CANCELLATO
%VIVIANA 

Creativity is one of the driving forces of human kind. Learning helps to catch up with current knowledge and current understanding of the world. Instead, creativity allows us to break current understanding to envision new ideas, which may revolutionize entire fields of knowledge. 
% V -> aggiunto us dopo allow

Scientific research offers a challenging environment where to explore and, eventually, learn regularities underlying the creative process: the creative process is here formal and documented in texts. In fact, scientific research is a creative act in the formal settings of the scientific method. ``Standing on the shoulders of giants'', scientists have new ideas, which aim to go beyond current understanding of the world. These ideas stem from existing knowledge and from creative thinking of a
group of scientists. Moreover, scientists are forced to document their creative process by publishing papers. In these papers, the creative process is somehow described. The background knowledge is declared, that is, references are provided and findings are described.  Creativity in scientific research is tracked, documented. 

In scientific research, the creative process can be seen as a text-to-text process. In fact, papers containing novel ideas are texts and are ''produced from'' referred papers, which are texts. This is a tremendous opportunity to study how to replicate a well-defined creative process.
% V -> aggiunto well-defined

Deep neural networks are giving the impression that it is possible to attack and resolve sequence-to-sequence tasks and, hence, text-to-text tasks. Sequence-to-sequence (Seq2Seq) neural networks are used to learn conversational agents from dialog \cite{Ghazvininejad2017AKN}. In this case, Seq2Seq NNs have to learn the relation between a stimulus utterance and a response utterance. There is not a direct relation between the two utterances but these systems have positive results. Moreover, Seq2Seq NNs are used for abstractive summarization \cite{rush-etal-2015-neural,chopra-etal-2016-abstractive,nallapati-etal-2016-abstractive,see-etal-2017-get}. Abstractive summarization is a text to text task and it is a summarization that may use different words with respect to those in target documents. Finally, encoder-decoder architectures are used to generate textual captions of images and, also, medical reports from medical images. Text-to-text tasks seem to be in the possibility of today's neural networks.
%trovare la bibliografia, ma ricordarsi cjhe la bibliografia totale non deve eccedere 1 pagina altrimenti lo cassano
In this paper, we dare to introduce a novel, challenging task of Generating Abstracts of Scientific Papers from abstracts of cited papers (GASP), we propose the GASP corpus and we experiment with the GASP corpus using three vanilla systems. The GASP task is a reduced version of the scientific creative process documented in papers. We define it a text-to-text task as follows: by having abstracts of cited papers, produce the abstract of the current paper. To foster research in this novel, challenging task, we prepared a dataset by using services where that solve the problem of copyright and, hence, the dataset is public available with its standard split.  Finally, we experimented with three vanilla summarization systems to start the analysis of the complexity of the GASP task. 
% V -> aggiunto a dopo The GASP task is
% V -> DOMANDA: non parsifico la frase ''where that solve the problem of copyright and, hence, the dataset is public available with its standard split''

The major contributions of this paper are:
\begin{itemize}
    \item the GASP task - a novel, challenging task capturing scientific creativity;
    \item the GASP corpus - a corpus where to test text-to-text systems for this novel, challenging task;
    \item the initial analysis of three vanilla summarization systems on the GASP task.
\end{itemize}
%qui io metterei almeno un accenno dei risultati anche se preliminari della sperimentazione  eche forniamo una baseline
The rest of the paper is organized as follows. Section \ref{sec:back} reports on the background and related work. Section \ref{sec:ciccia} formally define the GASP task and describes the collected corpus. Section \ref{sec:vanilla} shortly describes the three vanilla systems used in the experiments. Section \ref{sec:experiments} reports on the experiments with the vanilla systems on the GASP corpus. Finally, Section \ref{sec:conclusions} draws some conclusions and envisages future activities.  

\section{Background and related work}
\label{sec:back}
%\writers{FMZ}{Roberto, Giorgio}

Automatic generation of scientific papers has been attacked in the past and it has been seen as a way to test the scientific validity of specific conferences. Systems like SCIgen\footnote{\url{https://pdos.csail.mit.edu/archive/scigen/}} generate random papers by using a probabilistic context-free grammar in a generative way. Clearly, this is a very different case with respect to the GASP task. 

Generating Abstracts of Scientific Papers from abstracts of cited papers (GASP) seems to be strongly related to abstractive multi-document text summarization \cite{nallapati-etal-2016-abstractive,gulcehre-etal-2016-pointing,DBLP:conf/iclr/PaulusXS18}.  Indeed, abstractive text summarization has been already used to generate the \emph{related work} section of scientific papers \cite{hu-wan-2014-automatic,c79428a85c4e452cbadbb8aed1c7456c} or to automatically generate survey papers \cite{10.1007/978-3-030-18576-3_5}. 

GASP has similarities but also important differences with respect to abstractive summarization. In fact, in abstractive summarization target and, hence, generated summaries contain new phrases with respect to source documents: this is similar to our GASP task, where target abstracts generally may contain novel sentences, which are not in abstracts of cited papers. However, GASP is not only a summarization task. Scientific papers and, consequently, abstracts of scientific papers must contain a degree of novelty with respect to cited papers. Hence, GASP is definitely a novel, intriguing task, which aims to investigate scientific creativity.  

Although different, abstractive multi-document text summarization can help to envisage how to use neural network models and how to evaluate different systems. Neural networks have gained a lot of attention in automatic abstractive summarization \cite{DBLP:journals/corr/abs-1804-04589} since the task is seen as a sequence-to-sequence NLP task. The baseline system for encoding source texts seems to be a bag-of-word encoder \cite{rush-etal-2015-neural}, while systems for automatic abstractive summarization use convolution neural networks (CNN) \cite{rush-etal-2015-neural,chopra-etal-2016-abstractive} and recurrent neural networks  (RNN) \cite{nallapati-etal-2016-abstractive,see-etal-2017-get}. 

Evaluating systems aimed to produce text is a very difficult and debated problem, both in Machine Translation (MT) and Automatic Summarization (AS). The risk is to refer to measures that penalize good behavior of systems. In MT or AS, it is important that systems produce output that semantically cover what is in the reference: if words are not exactly the same, it is not a major issue. Hence, the best way to evaluate systems is by using human assessment \cite{coughlin2003correlating}. However, this evaluation methodology is extremely expensive. 

Many reference-based metrics have been proposed and largely used for evaluating MT and AS. Among all these metrics, the most common are BLUE \cite{papineni2002bleu}, ROUGE \cite{lin-2004-rouge} and METEOR \cite{banerjee2005meteor}. These measures are different as based on different principles. BLEU \cite{papineni2002bleu} is  precision-based: it counts how many $n$-grams of the generated text are in the ground-truth reference(s), with a penalization factor for repeated $n$-grams. Generated texts repeating $n$-grams receive a lower score. However, BLEU gives better scores to short generated texts as, for any precision-based measure, the less systems say the better these systems are evaluated. On the other side, there is ROUGE \cite{lin-2004-rouge}, which is recall-based. In this case, the metric counts the number of $n$-grams in references that are covered b  y generated text. There is also a variant that uses the size of the longest subsequence in common between references and generated texts. Clearly, as any recall based metric, ROUGE favors long generated texts even if these text may contain irrelevant information. To mitigate the problems of fully precision-oriented and fully recall-oriented measures, METEOR \cite{banerjee2005meteor} has been proposed as the harmonic mean between precision and recall over unigrams. In METEOR, references and generated texts are aligned before the computation of recall and precision. Then, this measure uses also a penalizing factor, which detects fragmentation by counting unigrams that are close in references but are far in generated texts. 

The GASP task we propose here has evaluation issues as those in Machine Translation and Automatic Summarization. Hence, we should be careful to choose the correct metrics, in order not to penalize good GASP task solutions.

\begin{figure*}
\begin{center}
\begin{subfigure}{.31\textwidth}
  \centering
  \includegraphics[width=0.99\textwidth]{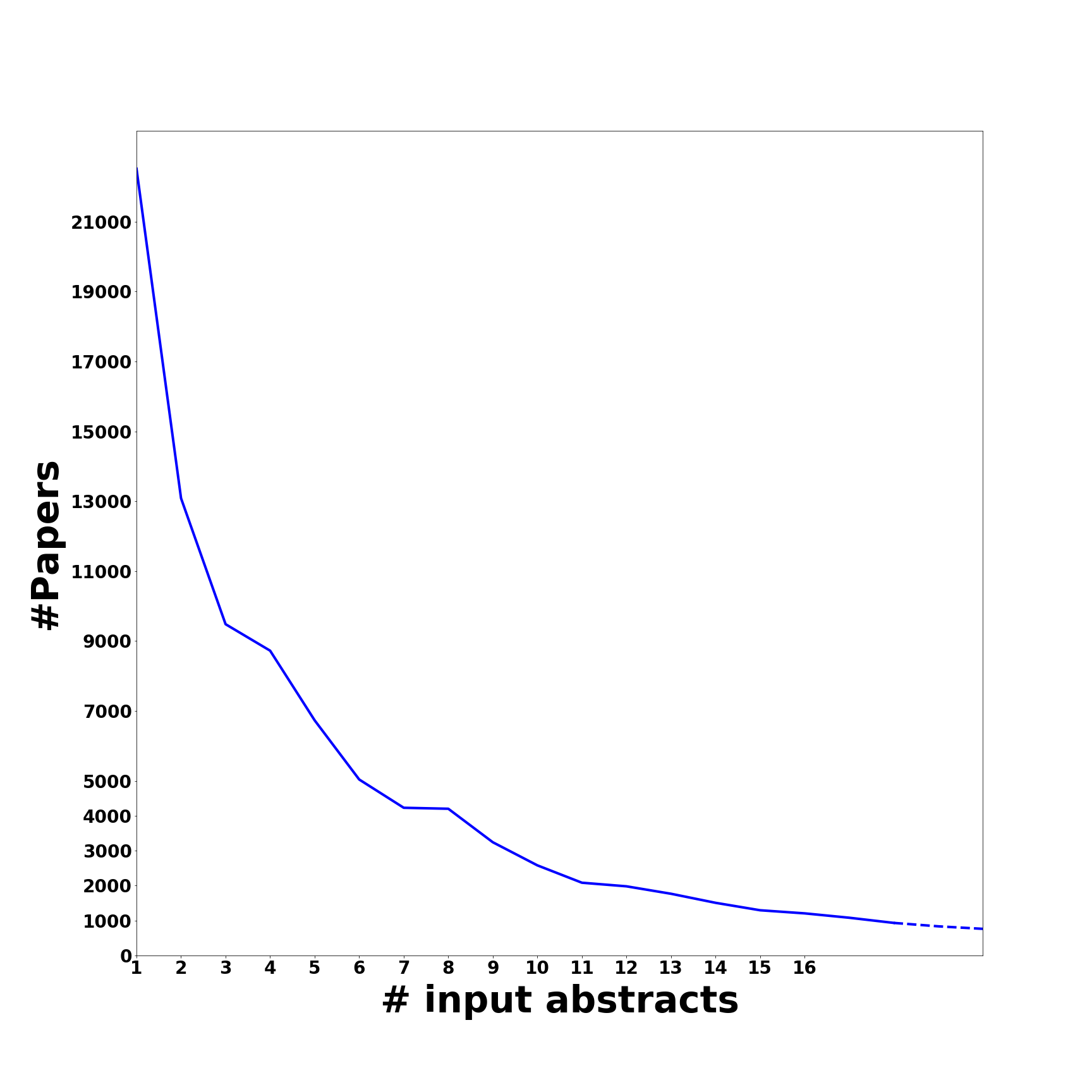}
  \caption{Source abstracts per target papers}
  \label{fig:sfig1}
\end{subfigure}%
\begin{subfigure}{.31\textwidth}
  \centering
  \includegraphics[width=0.99\textwidth]{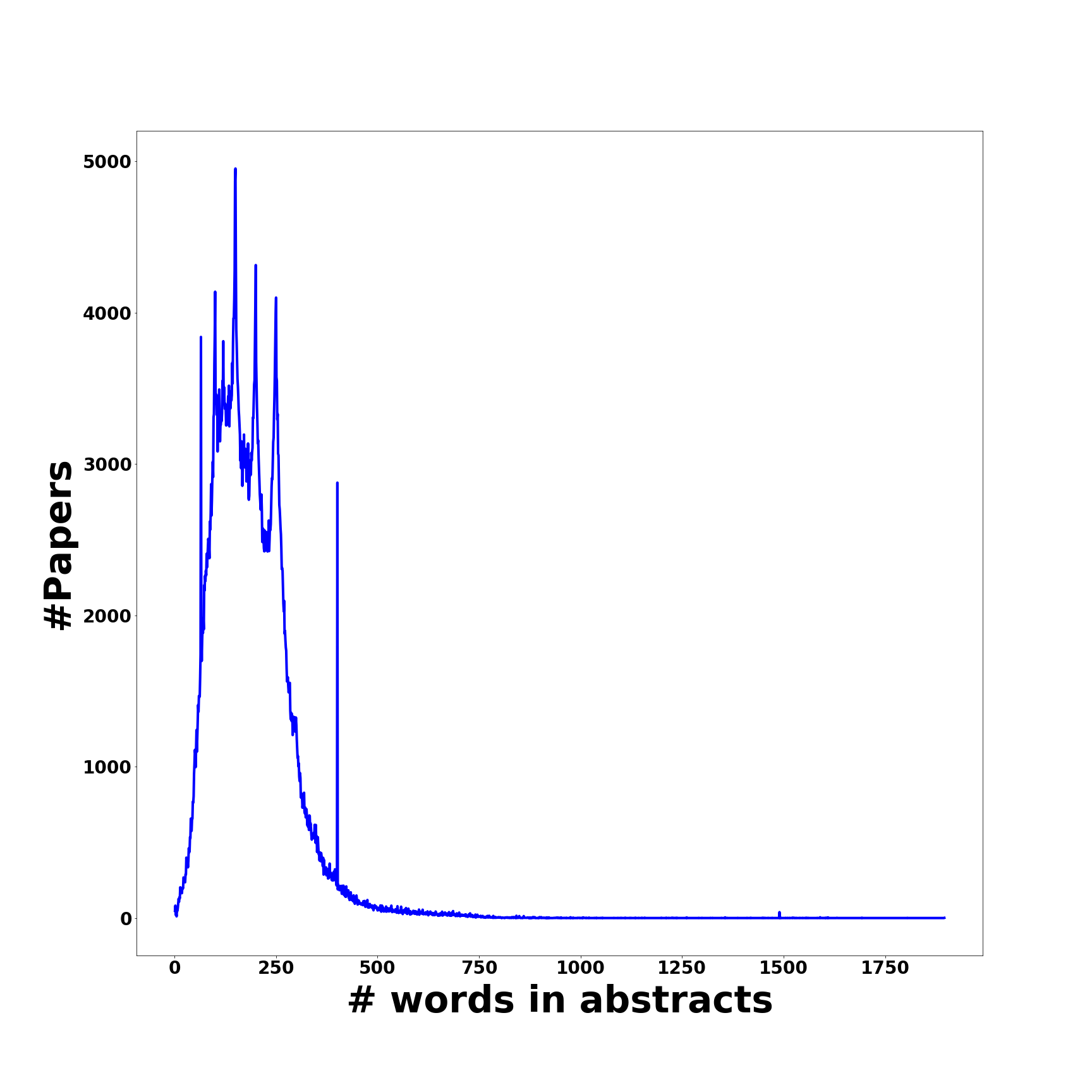}
  \caption{Words in papers}
  \label{fig:sfig3}
\end{subfigure}
\begin{subfigure}{.31\textwidth}
  \centering
  \includegraphics[width=0.99\textwidth]{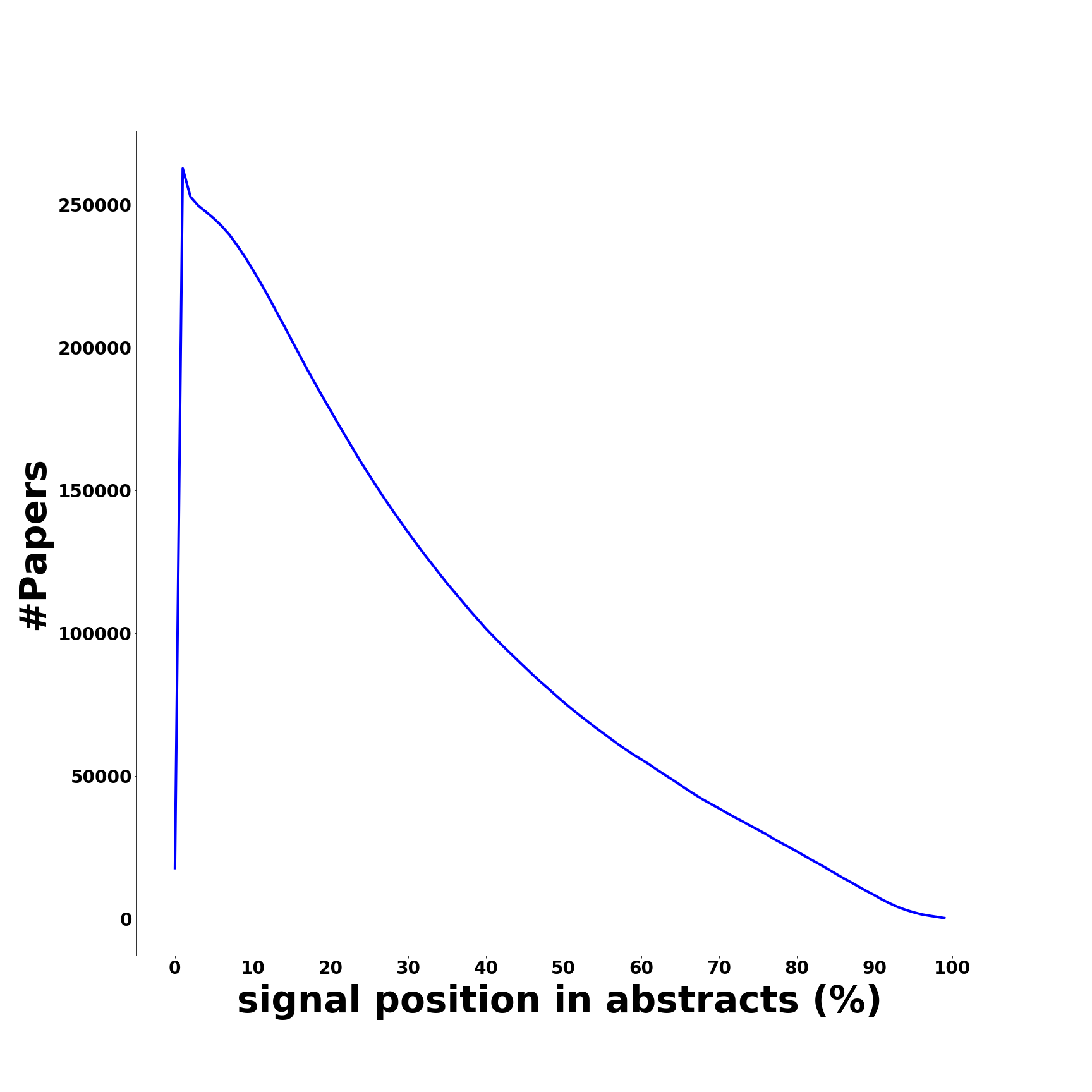}
  \caption{Starting contribution signal 
  %position in the length of abstracts
  }
  \label{fig:sfig4}
\end{subfigure}

\caption{The GASP task: corpus facts and statistics}
\label{fig:fig}
\end{center}
\end{figure*}

\begin{comment}
    \begin{subfigure}{.24\textwidth}
      \centering
      \includegraphics[width=0.99\textwidth]{papers_per_year.png}
      \caption{Distribution of target papers per year}
      \label{fig:sfig2}
    \end{subfigure}
\end{comment}

\section{Capturing Scientific Creativity in a Formal Task and in a Shared Dataset}
\label{sec:ciccia}
\writers{Fabio}{Roberto, Giorgio}

%The possibility of capturing scientific creativity in a formal task has always been there. The contribution of this paper is that we have recognized its existence. Then, we have plenty of opportunities to build up a dataset for the novel, challenging GASP task since we aim to focus on metadata of articles.
%There are millions of records for paper metadata, which are publicly available. Hence, we have a huge opportunity to build the GASP corpus and this opportunity has never been exploited, to the best of our knowledge. 
The following sections provide a thorough definition of the GASP task, the procedure and the data sources for building the GASP corpus and the corpus itself.

\subsection{GASP! The Task}

The GASP task aims to capture, at least in part, the creativity used to produce novel ideas. For this purpose, GASP refers to scientific papers, which describe the creation of novel ideas by using existing ideas, that is referenced papers. %Since we aim to devise a task that has a possibility to be solved, 
For the sake of simplicity, the definition of the GASP task is built upon abstracts of papers and not on full papers.  

The GASP task is aimed to producing the abstract of a paper - \textbf{the target abstract} - given the abstract of the set of referred papers - \textbf{the source abstracts} - which we may assume have been inspirational for the idea in the target paper. Hence, the formal definition is the following:
\[
[\{s_1,s_2,\ldots,s_n\}, t]
\]
where $t$ is the \textbf{target abstract} and $s_1,\ldots, s_n$ is the set of \textbf{source abstracts} of the inspirational papers. The task is then to produce $t$ by reading all $s_i$. Both $t$ and $s_i$ are to be intended as sequences of words.
The following is a example of $[\{s_1,s_2,s_3\}, t]$, which shows what the GASP task aims to be:
\begin{center}
\begin{tabular}{cp{5cm}}
    $t$ :  & United States prisons release more than 600,000 individuals each year. Within three years of release, 50 percent of released prisoners are back in prison. [...] \textbf{I} find that inmates who participate in work release have better post-prison employment outcomes.\\
    \hline
     $s_1$ : & Employment programs for disadvantaged male youth have been suggested as a possible new weapon in America's War on Drugs. \textbf{In this paper} panel data at the neighborhood level are used to investigate [..]  \\
     $s_2$ : & \textbf{This paper} empirically assesses the wage effects of the Job Corps program, one of the largest federally-funded job training programs in the United States. [...] \\
     $s_3$ : & \textbf{We} estimate the post-release economic effects of participation in prison-based General Educational Development (GED) programs using a panel of earnings records  [...]
\end{tabular}
\end{center}
By using the sequences of words in $s_1$, $s_2$ and $s_3$, systems are expected to produce $t$=\emph{United States prisons release more [...]}. As shown in the example, abstracts have an important feature. Abstracts are divided in two halves: a sort of introduction and the discussion of the novel idea. These two parts are split by signal word. In the above example these words are represented in bold text.

\subsection{Corpus Statistics and Facts}

Hence, the GASP task is a text-to-text task, which is similar to many sequence-to-sequence tasks. However, the GASP task is clearly different with respect to other text-to-text tasks as: (1) required outputs, that is target abstracts, and inputs, that is, source abstracts, are generally longer than in other tasks; (2) large part of target abstracts is novel information generally introduced by signal words. For these features, GASP is extremely difficult.

\subsection{Data Sources and Extraction Procedures}
\writers{Andrea}{Fabio}

There are millions of records for paper metadata, which are publicly available. Hence, we have a huge opportunity to build the GASP corpus.

Among all the data sources, we collected the GASP Corpus by using Semantic Scholar\footnote{\url{https://www.semanticscholar.org/}}. There are at least two good reasons for using Semantic Scholar for the GASP Corpus: (1) Semantic Scholar is already open to publicly share meta data of papers, since Semantic Scholar Open Research Corpus \cite{ammar:18} is already available; (2) Semantic Scholar offers APIs to eventually reconstruct the corpus or fill missing information by asking directly at the live servers. Hence, Semantic Scholar gives the opportunity to build an open corpus and gives a guarantee that GASP can be shared\footnote{The corpus can be downloaded at \url{https://github.com/ART-Group-it/GASP}}.

Then, the GASP Corpus has been then build by selecting papers from the Semantic Scholar Open Research Corpus (SSORC) \cite{ammar:18}. In SSORC, papers are represented with many metadata: for GASP, we are interested to the following fields: \emph{abstract} and \emph{outCitations}, which are the referred papers captured in the corpus. We selected 120,000 papers that have the outCitations with a least one paper. We aimed to build up a corpus with a training, testing and validation set of, respectively, 100.000, 10.000, and 10.000 instances.
Hence, for each paper in the 120,000, we built an instance line as follows:
$t$ is directly taken from the metadata of the paper; $s1$, ..., $s_n$ are extracted by reading the list in \emph{outCitations} and recovering abstracts from inside SSORC or by using the web APIs of Semantic Scholar \footnote{https://api.semanticscholar.org} if at least $3/4$ of  \emph{outCitations} list have been already covered.

In order to have the possibility to explain results of text-to-text neural networks on the GASP task, we here examine some facts of the GASP Corpus. In fact, this corpus has some particularities which will challenge these systems based on NNs. 

%\textbf{TODO: change peaks}
First, the length of most target abstracts is below 400 words with peaks at 100, 150, 200, 250, 300 and 400 justified by the existence of some limits in the paper format (see Fig. \ref{fig:sfig3}). There is a very long tail with abstract reaching the length of 1,400 words. This length of output texts and input texts are very challenging for end-to-end systems.  

Second, the number of source abstracts per target paper is extremely small. In fact, most of the target papers have only up to four source abstracts (see Fig. \ref{fig:sfig1}). 
%We performed some statistical analysis on the GASP corpus.
%The number of input abstracts per each paper is depicted in the plot \ref{fig:sfig1}. Unfortunately, most of the papers in the corpus have only one to four input abstracts. 
This fact depends on three different sources of problems: 1) target papers have really up to four cited papers; 2) the procedure to analyze citations using SSORC failed to cover some  citation references; 3) our extraction procedure failed to recover abstracts of some referred paper even if metadata exist for these cited papers.
Having few input abstracts can be a potential problem. In fact, the task could be really hard to solve and a system could be unable to generalize correctly. 
%In figure \ref{fig:sfig3} we analyzed the word length of abstract and we can conclude that is sufficient for a system in order to learn significant information.

Third, target papers relies on different sets of source papers. In fact, given the subset of target papers with only one source paper, the ratio of unique source papers with respect to all source papers is 0.89. This ratio is around 1 for all the target papers that have more than one source papers. This means that nearly each set of input papers is unique. Hence, it is guaranteed that given a set of source papers, there is only one single target paper that is produced. The GASP dataset is then coherent as only one idea is derived from  a set of inspiring ideas.  

Finally, one other important fact is the distribution of pre-existing knowledge with respect to novel ideas in target and source abstracts. This is extremely important as pre-existing knowledge should be easier to reproduce with respect to novel knowledge. Abstracts are generally organized in two parts. A first part describes the problem and current solutions. A second part describes the novel idea proposed in the paper. There is a clear separation between the two parts as the second part is introduced by a signal phrase. 
We analyzed the GASP Corpus by using the following signal phrases \textit{'I', 'We', 'in this paper', 'in this work', 'our approach', 'our work', 'this paper', 'this work', 'this study'}. 
Abstracts in the GASP Corpus are unbalanced since most of these abstract report on the novelty introduced by the paper. In fact, a large amount of abstracts have the signal phrase before the 30\% of their length (see Fig. \ref{fig:sfig4}).  
Hence, the GASP task is extremely challenging as most of the content in abstracts is novel and cannot result from a simple summarization of source abstracts.  

%e analyzed input-papers abstracts to look for indicators of beginning of the authors' contribution in the length of the abstract. For each paper, we calculated the position in the abstract of the first occurrence of the signal normalized by the total length of the abstract. We decided to focus on the following starting contribution signals: . Results are shown in figure \ref{fig:sfig4}. From the plot we can conclude that in most of the papers the authors' contribution start in the first half of the abstract, in this case we expect a system to start to generate the novelty quite soon in the abstract.

%We evaluated also how many unique input abstracts lists there were among corpus entries. We found that for corpus entries with only one input abstract, $0.89\%$ of entries were unique. For the other entries, with more than one input abstract, the percentage is $\sim1$

\begin{comment}
\begin{center}
\begin{table}[!h]
\begin{tabular}{|l|l|l|ll}
\cline{1-3}
          & $\mu$  & $\sigma^2$ &  &  \\ \cline{1-3}
\# words  & 192   & 9060     &  &  \\ \cline{1-3}
\# abstracts & 7     & 61       &  &  \\ \cline{1-3}
coverage  & 0.995 & 0.001    &  &  \\ \cline{1-3}
\end{tabular}
\end{table}

\end{center}
\end{comment}

\section{Vanilla Systems}
\label{sec:vanilla}
\writers{Andrea}{Danilo}

We considered the GASP task as an abstractive summarization task. In a text summarization task, input is composed by pairs $(X,Y)$ where the input $X$ is composed by a series of tokens $w_1,\ldots, w_n$ and the output $Y$ is composed by a series of tokens $y_1,\ldots, y_m$, with $m<<n$, in general. 
However, in the GASP Corpus input $X$ is composed by a variable-lenght series of papers $s_1,s_2,\ldots,s_n$ each of which is composed of a series of tokens $w_1,\ldots,w_j$. In our systems we decided to concatenate the sequence of tokens of each paper $s_i$ to reduce the problem to a general text summarization problem. Each token was given to the systems as it is without a particular preprocessing procedure.

% The procedure on how the exact concatenation happend are explained in the respective system section.
%each element $[\{s_1,s_2,\ldots,s_n\}, t]$ of the corpus in this way: $ w_i^{s_1},\ldots,sep, w_i^{s_2}, \ldots, sep,  w_i^{s_n}$
In the next subsections, we proceed by describing  each system in detail.

%We built several systems and evaluated the performance of each system on the LAST corpus.

\begin{center}
\begin{table*}
\begin{small}
\begin{tabular}{p{5cm}|p{5cm}|p{5cm}}
\emph{Extractive Summarization} & \emph{Abstractive Summarization} \\
\textbf{TextRank} & \textbf{BiLSTM} & \textbf{Gold Target Abstract}  \\ 
\hline
\textit{In this paper, we use a unique set of individual-level proposition voting data from Los Angeles County that allows us to estimate the distribution of voter preferences, including the mean, median, and variance (heterogeneity), for various subsets of voters in each of 55 State Senate, State Assembly, and U.S} & \textit{In this paper we examine the impact of citizen on the distribution of voter preferences in the context of the plating roll We show that there is a strong relationship between the preferences of the median and the preferences of the median We show that there is a relationship between} & Contrary to popular opinion we find evidence that the views of residents of both higher and lower income neighborhoods are represented by their legislators. Analyzing the voting behavior of California state legislators on 77 proposals on which both the legislature and the public cast ballots we find first that [...]  \\ \hline

%\textit{In this paper, we use a unique set of individual-level proposition voting data from Los Angeles County that allows us to estimate the distribution of voter preferences, including the mean, median, and variance (heterogeneity), for various subsets of voters in each of 55 State Senate, State Assembly, and U.S. House districts in the county.\nRegression analyses reveal that legislators’ roll call votes are more closely related to district mean preferences in homogeneous districts than they are in heterogeneous districts.[...]} & \textit{In this paper we examine the impact of citizen on the distribution of voter preferences in the context of the plating roll We show that there is a strong relationship between the preferences of the median and the preferences of the median We show that there is a relationship between} & \textit{System3 output} & "Contrary to popular opinion  we find evidence that the views of residents of both higher and lower income neighborhoods are represented by their legislators. Analyzing the voting behavior of California state legislators on 77 proposals on which both the legislature and the public cast ballots  we find first that the opinions of higher and lower income voters within a district are highly correlated and thus it is impossible to represent the views of one group and not also represent the views of the other. [...]  \\ \hline
\textit{The TIPSTER Text Summarization Evaluation (SUMMAC) has established definitively that automatic text summarization is very effective in relevance assessment tasks. 1 Introduction In May 1998, the U.S. government completed the TIPSTER Text Summarization Evaluation (SUMMAC), which was the first large-scale, developer-independent evaluation of automatic text summarization systems. 1.1 Text Summarization Text summarization} & \textit{In this paper we propose a new method for text summarization in the context of text summarization in the context of text summarization systems. The method is based on a set of text summarization tasks that are used in the context of text summarization tasks. The method is based on} & This paper describes a framework for multidocument summarization which combines three premises: coherent themes can be identified reliably; highly representative themes  running across subsets of the document collection  can function as multi-document summary surrogates; and effective end-use of such themes should be facilitated by a visualization environment which clarifies [...] \\ \hline

%\textit{System 1} & \textit{System2} & \textit{System3} & \textit{Creativity is one of the driving forces of human kind as it allows to break current understanding to envision new ideas, which may revolutionize entire fields of knowledge. Scientific research offers a challenging environment where to learn creativity. In fact, scientific research is a creative act in the formal settings of the scientific method and this creative act is describe in articles. In this paper, we aim to study whether deep learning models can be creative and, hence, we introduce a novel, challenging task: the GASP task - Generate Abstracts of Scientific Papers from Abstracts of Cited Papers. The GASP task is defined as follows: by having abstracts of cited papers, produce the abstract of the current paper. We propose a dataset and a standard split where to test different sequence-to-sequence systems. } \\

\end{tabular}
\end{small}
\caption[2]{Sample System Outputs and expected target abstracts}
\label{tab:tb2}
\end{table*}
\end{center}

\subsection{TextRank}
%baseline bag-of-word as in \cite{rush-etal-2015-neural}
a
As the name implies, this system was built using the TextRank algorithm \cite{textrank}. Input abstracts were concatenated as a whole document separated by the token \textit{\textbackslash n} and the result was the input to the TextRank algorithm.
%We used the TextRank implementation from \cite{textrank2} limiting the word lenght of the summary to 50 words.
Clearly the TextRank algorithm is not particularly suited  for the GASP task due to the fact that it is an extractive text summarization algorithm. This means that words, phrases, sentences of the output abstract are selected from source abstracts, hence this algorithm, and in general any  extractive text summarization algorithm, is not able to capture the creative-generative process behind authors' intent which is the aim of the GASP task. However, we decided  to use this system as a baseline anyway, being aware of the limitations it has.

\subsection{BiLSTM}
BiLSTM is a standard sequence-to-sequence system used for text summarization trained with maximum likelihood loss function for the sequence labeling problem. As for the TextRank system, the input sequence is composed of input abstracts concatenated as a whole document and separated by the tag \textit{[SEP]} to delimit the end of an abstract and the beginning of the next one. The output sequence is simply the output abstract.
%Each input and output word $w_i$ is mapped to a vector representation realized concatenating glove word embedding \cite{pennington2014glove} with Elmo contextual word embedding\cite{elmo} fine-tuned for the task.
%dimension word embedding glove
The system was built with a bidirectional LSTM \cite{lstm} with attention \cite{bahdanau2014neural} with copy mechanism \cite{point} that allow it to copy input words to the output.
%. The model use also a copy mechanism \cite{point} that allow it to copy input words to the output.
%Input was limited to $500$ tokens, while output was truncated to $50$ tokens. The system was trained with Adagrad \cite{adagrad} with a learning rate of $0.15$, an initial accumulator value of $0.1$, batch of $32$ for $20$ epoch.

\begin{comment}
\subsection{Transformer}
Transformer is a sequence-to-sequence system using transformer architecture proposed by \cite{DBLP:journals/corr/VaswaniSPUJGKP17}. We used OpenNMT with the configuration for text summarization \cite{gehrmann2018bottom}. The input was processed as explained for the BiLSTM system.
\end{comment}

%\newpage

%\textbf{TODO: descrivere come verrà messo a disposizione il corpus}
\begin{center}
\begin{table*}
\begin{small}
\begin{tabular}{l|l|l|l|l|l|l|l|l|l|l|l|l|l}
    & \emph{R2 Recall} & R2 Precision & R2 F1  & R1 Recall & R1 Precision & R1 F1 & RL Recall & RL Precision & RL F1\\ \hline
TextRank & 0.053 & 0.018 & 0.024 & 0.325 & 0.118 & 0.161 & 0.207 & 0.074 & 0.101  \\ \hline
BiLSTM & 0.019 & 0.058 & 0.027  & 0.109 & 0.334 & 0.154 & 0.077 & 0.238 & 0.108 \\ \hline
%Transformer & nan & nan & nan  & nan & nan & nan & nan & nan & na  \\ 
\end{tabular}
\end{small}
\caption[2]{Rouge performance of TextRank and BiLSTM on the testing of the GASP Corpus}
\label{tb:1}
\end{table*}

\end{center}

\section{Experiments}
\label{sec:experiments}
\writers{Andrea}{Fabio, Danilo}
We performed some initial experiments to valuate the complexity of the GASP task. Hence, we experimented with the GASP corpus and with the vanilla systems presented in the previous section. After a description of the experimental details in Sec. \ref{sec:exp_set_up}, we discuss results in Sec. \ref{sec:results}.

\subsection{Experimental Set-up}
\label{sec:exp_set_up}

To have the possibility to, at least, perform the experiments, we constrained the GASP task: we cut target abstracts to 50 words. This is needed for the computational cost of the abstractive summarization models we used. Analyzing this reduced version of the GASP corpus is still interesting to start to explore the complexity of the task. 

In the experiments, we used the following implementations of the above vanilla systems. 
For the TextRank system, we used the implementation from \cite{textrank2}\footnote{https://github.com/summanlp/textrank}. For comparison purposes and not for computational limitations, we constrained the output of the system to abstract up to  50 words. This limits allow to compare resuls with the other system we used. 
For the abstractive summarization based on BiLSTM, we used OpenNMT \cite{opennmt} with the configuration for abstractive text summarization \cite{gehrmann2018bottom}. The system implements a bidirectional LSTM of $256$ units. For computational purposes, we constrained the input to $500$ words. This means that part of the source abstracts are not considered during the training. 
Target, output abstracts are already constrained to be less than $50$ words. 
As optimizer, we used Adagrad \cite{adagrad} with a learning rate of $0.15$, an initial accumulator value of $0.1$. The size of the batch is $32$ and we run  $20$ epochs. To allow replicability of the experiments, we released the configuration file\footnote{https://github.com/ART-Group-it/GASP}.

To compare systems, we used the python implementation easy-rouge\footnote{https://github.com/pltrdy/rouge} of the Rouge \cite{lin-2004-rouge} evaluation metric since these measure is widely used to evaluate  text summarization systems. In particular we evaluate Recall, Precision and F1 measure for the metrics Rouge-1 (R1), Rouge-2 (R2) and Rouge-L (RL).

\subsection{Results and Discussion}
\label{sec:results}

%TODO: talk about \ref{tab:tb2}

Both the extractive and the abstractive systems produce some reasonable text for target abstracts. For example, line 1  of Table \ref{tab:tb2}  shows the extractive summarization that describes the relation between preferences of voters and voting data. This is similar to what is described in the gold target abstract. Whereas, line 2 shows the absractive summarization system BiLSTM that produces an abstract related to summarization and the gold target abstract is in fact related to multidocument summarization. 
From these examples, it seems that BiLSTM system fails to get the creative-generative intent behind the authors, but it is able to get the topic of target abstracts and give same coherent text around the topic.

%Results of systems performances are shown in table \ref{tb:1} while two selected outputs are shown in table \ref{tab:tb2}. 

In general, the extractive text summarization algorithm TextRank tends to have and higher recall and a lower precision respect to the other system (see Table \ref{tb:1}). This happens because the TextRank algorithm reuse words ans sentences in source abstracts to match the fixed 50 words length of the summary.
 In contrast, the abstractive BiLSTM system tends to be more precise losing points in recall. However, some BiLSTM outpus are really extremely odd.
For example, mostly for medical paper, the BiLSTM system performs poorly producing short texts like:  \textit{Clinical pulmonary a a a a a a a a a a}. This fact is due probably to topic distribution of papers in the training set that could be overcome with an extended version of the GASP corpus. 

All the systems performed very bad on the task. This confirms the  complexity of the GASP task. 
We think that the reason why all the systems performed in this way is strictly related to the nature of the task: trying to learn a statistical correlation between input papers and output paper is not enough to capture the creativity intent of the authors. We speculate that trying to solve the task is strictly related to a deep understanding of the generative process behind the writing process, hence we propose the GASP task as a way to build better machine learning models capable to grasp author's intent.

%\subsection{Discussion}
%\writers{Giorgio}{FAbio}

\section{Conclusions}
\label{sec:conclusions}
\writers{Fabio}{Viviana, Paola}

 To the best of our knowledge, this is the first paper that introduces the task of modeling scientific creativity. By proposing the task of generating abstracts of scientific papers from abstracts of cited papers, we opened an opportunity to build text-to-text systems attacking a task we don't want to solve. We picked the hanging GASP dataset, which  has been always in front of us, we delivered it and we started to analyze the performance of existing vanilla systems. Luckily, the timid results are still far from being satisfactory but show some encouraging directions of study. \emph{Alea iacta est}. We believe GASP poses an intriguing, difficult, and philosophically important challenge for the artificial intelligence field.

\bibliographystyle{acl_natbib}
\bibliography{last}

\end{document}